\documentclass[sigconf]{acmart}
\usepackage{multirow}
\usepackage{subfigure}

\usepackage{amsthm,amsmath}
\usepackage{mathrsfs}
\usepackage{enumerate}
\usepackage{graphicx}
\usepackage{booktabs}
\usepackage{multirow}
\usepackage{tabularx}

\newcommand\Tstrut{\rule{0pt}{2.6ex}}

\usepackage{algorithmic}
\usepackage[ruled,commentsnumbered,linesnumbered]{algorithm2e}
\AtBeginDocument{%
  \providecommand\BibTeX{{%
    \normalfont B\kern-0.5em{\scshape i\kern-0.25em b}\kern-0.8em\TeX}}}

\copyrightyear{2023}
\acmYear{2023}
\setcopyright{acmlicensed}\acmConference[KDD '23]{Proceedings of the 29th ACM SIGKDD Conference on Knowledge Discovery and Data Mining}{August 6--10, 2023}{Long Beach, CA, USA}
\acmBooktitle{Proceedings of the 29th ACM SIGKDD Conference on Knowledge Discovery and Data Mining (KDD '23), August 6--10, 2023, Long Beach, CA, USA}
\acmPrice{15.00}
\acmDOI{10.1145/3580305.3599486}
\acmISBN{979-8-4007-0103-0/23/08}

\title{Recognizing Unseen Objects via Multimodal Intensive Knowledge Graph Propagation}

\author{Likang Wu}
\authornote{Equal Contributions.}
\orcid{0000-0002-4929-8587}
\affiliation{%
  \institution{School of Computer Science and Technology, University of Science and Technology of China \& State Key Laboratory of Cognitive Intelligence}
  \city{Hefei}
  \state{Anhui}
  \country{China}
}
\email{wulk@mail.ustc.edu.cn}

\author{Zhi Li}
\authornotemark[1]
\affiliation{%
  \institution{Shenzhen International Graduate School, Tsinghua University}
  \city{Shenzhen}
  \state{Guangdong}
  \country{China}
}
\email{zhilizl@sz.tsinghua.edu.cn}

\author{Hongke Zhao}
\affiliation{%
  \institution{College of Management and Economics, Tianjin University}
  \city{Tianjin}
  \country{China}
}
\email{hongke@tju.edu.cn}

\author{Zhefeng Wang}
\affiliation{%
  \institution{Huawei Cloud}
  \city{Hangzhou}
  \state{Zhejiang}
  \country{China}
}
\email{wangzhefeng@huawei.com}

\author{Qi Liu}
\affiliation{%
  \institution{School of Computer Science and Technology, University of Science and Technology of China \& State Key Laboratory of Cognitive Intelligence}
  \city{Hefei}
  \state{Anhui}
  \country{China}
}
\email{qiliuql@ustc.edu.cn}

\author{Baoxing Huai}
\affiliation{%
  \institution{Huawei Cloud}
  \city{Hangzhou}
  \state{Zhejiang}
  \country{China}
}
\email{huaibaoxing@huawei.com}

\author{Nicholas Jing Yuan}
\affiliation{%
  \institution{Huawei Cloud}
  \city{Hangzhou}
  \state{Zhejiang}
  \country{China}
}
\email{nicholas.jing.yuan@gmail.com}

\author{Enhong Chen}
\authornote{Corresponding Author.}
\affiliation{%
  \institution{School of Computer Science and Technology, University of Science and Technology of China \& State Key Laboratory of Cognitive Intelligence}
  \city{Hefei}
  \state{Anhui}
  \country{China}
}
\email{cheneh@ustc.edu.cn}



\begin{document}

%%
%% The abstract is a short summary of the work to be presented in the
%% article.
\begin{abstract}
Zero-Shot Learning (ZSL), which aims at automatically recognizing unseen objects, is a promising learning paradigm to understand new real-world knowledge for machines continuously. Recently, the Knowledge Graph (KG) has been proven as an effective scheme for handling the zero-shot task with large-scale and non-attribute data. Prior studies always embed relationships of seen and unseen objects into visual information from existing knowledge graphs to promote the cognitive ability of the unseen data. Actually, real-world knowledge is naturally formed by multimodal facts. Compared with ordinary structural knowledge from a graph perspective, multimodal KG can provide cognitive systems with fine-grained knowledge. For example, the text description and visual content can depict more critical details of a fact than only depending on knowledge triplets. Unfortunately, this multimodal fine-grained knowledge is largely unexploited due to the bottleneck of feature alignment between different modalities. To that end, we propose a multimodal intensive ZSL framework that matches regions of images with corresponding semantic embeddings via a designed dense attention module and self-calibration loss. It makes the semantic transfer process of our ZSL framework learns more differentiated knowledge between entities. Our model also gets rid of the performance limitation of only using rough global features. We conduct extensive experiments and evaluate our model on large-scale real-world data. The experimental results clearly demonstrate the effectiveness of the proposed model in standard zero-shot classification tasks.
\end{abstract}

\maketitle

\section{Introduction}
Zero-Shot Learning (ZSL), which utilizes prior knowledge to recognize unseen objects in the machine learning process~\cite{xian2018zero,wu2023survey}, becomes an important task to test the machine's cognitive ability. The popular approaches mainly focus on learning the semantic correlation between seen and unseen categories by constructing a unified space from manually annotated attributes ~\cite{liu2020attribute,xie2019attentive}. Although these methods have achieved promising performances in ZSL tasks, it is too difficult and time-consuming to annotate elaborate attributes for large-scale data in the real world. Recently, several researchers begin to cope with this challenge by incorporating the Knowledge Graphs (KG) to generate transferable knowledge and promote the cognitive ability of the ZSL models~\cite{wang2018zero,kampffmeyer2019rethinking}.

Although existing KG-based ZSL methods obtain great achievements in the task with large-scale data, they only exploit the structural knowledge to generate the representations of unseen objects. When identifying similar categories, they often fail to pay attention to key distinguishing features, resulting in classification errors.
% via embedding various objects in the KG. 
% Actually, real-world knowledge is naturally formed by multimodal facts, such as concept words, text descriptions, and referred images.
Actually, for this problem, multimodal knowledge (e.g. concept words, text descriptions, referred images, etc.) can help machines better understand the real world in a more fine-grained way and also promote the cognitive ability of ZSL. For example, Figure~\ref{fig:intro1} illustrates a toy example of multimodal knowledge graph in the ZSL task. Now ``Horse'' and ``White Tiger'' are seen classes for the machine while ``Zebra'' is unseen. Then, with structural knowledge, we teach the machine that ``Zebra'' is a subgenus of ``Horse''. With text descriptions, we can tell the machine that ``Zebra'' has black and white stripes same as ``White Tiger'' to help the machine adaptively capture the fine-grained semantics. Therefore, the machine can automatically recognize ``Zebra'' as a horse-like animal with black and white stripes after seeing the ``Horse'' and ``White Tiger'' images. From this example, we can conclude that multimodal knowledge can help machines promote cognitive ability for the real world furthermore achieve better ZSL performances.

% 挑战需要重新组织

Unfortunately, the exploration of multimodal knowledge in ZSL is largely limited by great challenges, which leads to the lack of mature paradigms for solving ZSL problems via multimodal knowledge graphs. First, it is hard to organize the multimodal knowledge and generate representations for the heterogeneous multimodal data. Second, the semantic associations of different modal knowledge are not easy to capture and construct. For example, the text descriptions have more fine-grained information (``black and white stripes'') than concept words as shown in Figure~\ref{fig:intro1}. Then, adaptively exploiting fine-grained information to enhance corresponding features in other modalities is significantly difficult, which leads to the dilemma of transferring multimodal knowledge from seen objects to unseen objects. And we know that knowledge transfer is the key to promoting ZSL learning.

To solve the dilemma mentioned above, in this paper, we present a focused study on the zero-shot recognition framework from a novel multimodal knowledge intensive perspective. Specifically, we propose a novel Fine-grained Graph Propagation (FGP) model to deeply exploit the fine-grained multimodal knowledge and transfer it from seen to unseen objects. We first introduce a semantic space to adaptively generate the various fine-grained semantic embeddings from the text descriptions for each concept. Then, a fine-grained visual knowledge learning mechanism is designed to prompt the image classifier to understand more knowledge from both visual and textual modalities, which adaptively maintains the fine-grained features aligned in the visual-semantic space simultaneously. Next, to transfer the multimodal knowledge from seen to unseen objects, we propose a Multi-facet Graph Convolutional Network (Multi-GCN) to learn unified fine-grained class representations for both the seen and unseen objects from the structural knowledge. Finally, we conduct sufficient experiments in real-world large-scale datasets. Experimental results and visualization cases clearly demonstrate the effectiveness of our proposed FGP. 
% The code is released online~\footnote{https://anonymous.4open.science/r/FGP-7144} without personal information.
In summary, the main contributions could be summarized as follows:

\begin{figure}[t]
\centering
\includegraphics[width=0.42\textwidth]{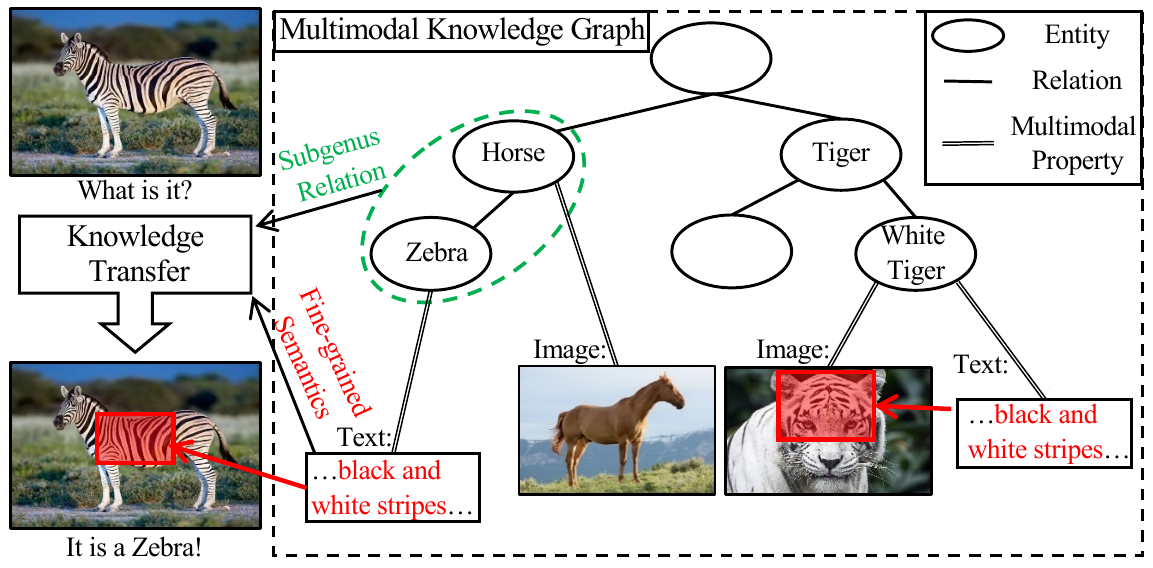}
% \vspace{-0.05in}
\caption{The knowledge transfer process of typical zero-shot learning with the fine-grained knowledge alignment mechanism on multimodal data.}
% \vspace{-0.2in}
\label{fig:intro1}
\end{figure}

\begin{itemize}
\item We present the zero-shot recognition framework from a novel multimodal knowledge intensive perspective.
\item We propose a Fine-grained Graph Propagation (FGP) model that deeply exploits the fine-grained multimodal knowledge and transfers it from seen to unseen objects. FGP breaks through the restriction that requires annotation attributes for ensuring visual-semantic alignment.
\item We conduct sufficient experiments on real-world large-scale datasets, and the experimental results show the superiority of our model compared with all baselines.
\end{itemize}

\section{Related Work}
The two most relevant technical points of this paper are zero-shot learning and graph neural networks. We introduce the related research work from these two aspects respectively.
\subsection{Zero-Shot Learning}
We will review zero-shot learning from two perspectives: embedding-based methods and KG-based methods. The motivation of embedding models is to represent all classes by trained vector representations that can be mapped to visual classifiers~\cite{changpinyo2017predicting,chen2018zero,fu2016semi,frome2013devise,zhang2022exemplar}. ~\cite{socher2013zero} trains two unsupervised neural networks for visual and text inputs and then learn a linear mapping between captured image representations and word embedding. After that, ~\cite{frome2013devise} proposes a novel framework DeViSE which trains a mapping relation from image to word embeddings via a ConvNet and a transformation layer. 
%By using the predicted embedding to perform a nearest neighbor search, DeViSE scales up the zero-shot recognition to thousands of classes. 
Instead of predicting the word embedding directly, ~\cite{norouzi2013zero} designs ConSE to construct the image embedding, which combines the image classification ConvNet and the word embedding model at the same time. In recent years, many studies implement the attention mechanism to learn fine-grained features to enhance the discriminative power of embedding~\cite{xie2019attentive,huynh2020fine,xie2020region,huynh2020compositional}.~\cite{huynh2020compositional} propose an interesting feature composition
model that captures attribute-based visual features from training instances and combines them to construct fine-grained features for unseen classes. The multimodal model~\cite{YuLiu2020ADM} incorporates a joint visual-attribute attention module and a multi-channel explanation module to try explainable zero-shot recognition.
Besides, with the help of ontology-based knowledge representation and semantic information, several recent studies explore richer and more competitive prior knowledge to model the inter-class relationship for knowledge transfer of ZSL~\cite{geng2021ontozsl,geng2022disentangled}.

To handle large non-attribute data with fine-grained features, we concentrate on the popular fashion of using multimodal knowledge graphs~\cite{chen2022msnea} (explicit knowledge representations) for this shortage. Researchers have proposed a unified pipeline on how to use knowledge graphs for ZSL object recognition~\cite{wang2018zero,kampffmeyer2019rethinking}. This idea has been early used in the work~\cite{wang2018zero,kampffmeyer2019rethinking}, where a graph convolutional neural network was trained to predict logistic regression classifiers on top of a pre-trained CNN to distinguish unseen classes. Later, HVEL learned the implicit knowledge and explicit knowledge in the hyperbolic space~\cite{liu2020hyperbolic}, which better captured the class hierarchy with fewer dimensions. ~\cite{wang2021zero} exploits multiple relationships among different categories of KG for zero-shot learning by employing graph convolutional representation and contrastive learning techniques.~\cite{xu2022vgse} visually divides a set of images from seen classes into clusters of local image regions according to their visual similarity, and further imposes their class discrimination and semantic relatedness. MGFN~\cite{wei2022semantic} develops a multi-granularity fusion network that integrates discriminative information from multiple GCN branches.
In other related applications, such as action recognition, DARK~\cite{luo2022disentangled} trains a factorized model by extracting disentangled feature representations for verbs and nouns and then predicting classification weights using relations in external knowledge graphs. However, there is no effective KG-based method utilizing multimodal features well, since it is challenging to ensure the alignment of information between different modalities without fine-grained annotations.
% But these models lack the ability of mining multimodal fine-grained knowledge from global images and text descriptions, which are the essential parts to promote machine’s  cognitive ability for ZSL task.
In this paper, we present a novel view to exploit this multimodal knowledge in the KG-based ZSL. 

% point our proposed multimodal solution FGP focuses on.

\subsection{Graph Neural Networks}
We just utilize graph neural networks as an efficient tool to handle our knowledge aggregation process in our work, so we only summarize some of the most classic representation learning methods. Graph neural networks (GNNs)~\cite{gori2005new,scarselli2008graph,wu2021learning}, especially graph convolutional networks \cite{henaff2015deep}, have activated many researchers' motivation in structured data modeling and analysis because of their theory simplicity and model efficiency~\cite{bronstein2017geometric,chiang2019cluster}. They break hard performance bottlenecks in many research areas, such as node classification~\cite{kipf2016semi,wu2023learning}, graph classification~\cite{defferrard2016convolutional}, and recommendation~\cite{zheng2023interaction,DBLP:journals/tkdd/QiuHW22}. The graph spectral theory is first used to derive a graph convolutional layer~\cite{henaff2015deep}. Then, the polynomial spectral filters are proposed to greatly reduce the computational cost~\cite{defferrard2016convolutional}. ~\cite{kipf2016semi} proposes the implementation of a linear filter to make further calculation simplification. Along with spectral methods, directly conducting graph convolution in the spatial domain is also investigated by so many studies~\cite{duvenaud2015convolutional,hamilton2017inductive}. Among them, graph attention network as a representative method has shown the greatest research potential~\cite{velivckovic2017graph}, which adaptively specifies weights to the neighbors of a node by attention mechanism~\cite{bahdanau2014neural,wu2020estimating}. 
For the GNN-based representation in zero-shot tasks,
~\cite{wang2021dpgn} proposes a GCN-based model named DGPN which follows the principles of locality and compositionality of zero-shot model
generalization.~\cite{LuLiu2020AttributePN} optimizes the learned attribute space for ZSL by training a propagation mechanism to refine the semantic attributes according to the neighbors and related classes. And~\cite{CaixiaYan2020SemanticsPreservingGP} explores
a graph construction approach to flexibly create useful category graphs
via leveraging diverse correlations between category nodes.
% Semantic Graph Convolutional Networks (SGCN)~\cite{wu2021learning} explores the implicit semantics by learning latent semantic-paths on graphs. In our model, GCN was applied to predict the last layer weight of visual feature extractor.

\begin{figure*}[t]
\centering
\includegraphics[width=0.85\textwidth]{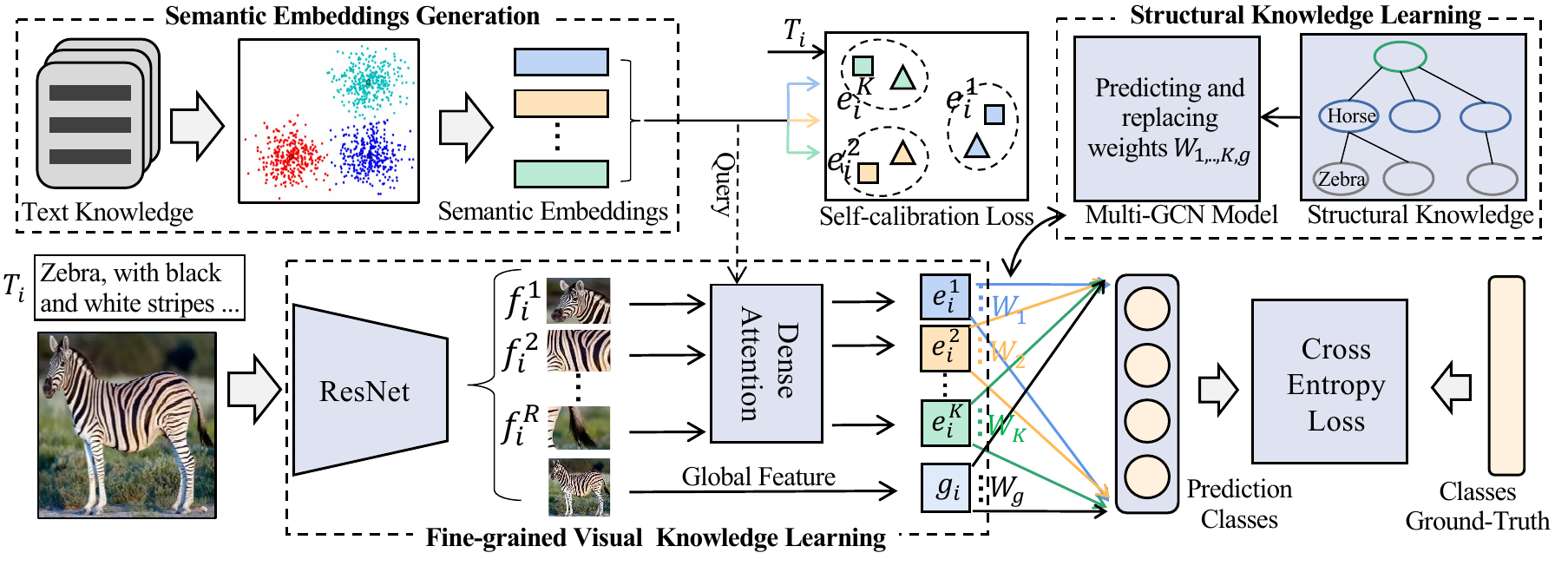} % Reduce the figure size so that it is slightly narrower than the column.
\caption{The overview of our zero-shot recognition framework, i.e.,  Fine-grained Graph Propagation (FGP). In detail, the two panels on the top indicate the semantic embedding generation process and structural knowledge learning, respectively. The bottom panel denotes our fine-grained knowledge extractor and classifier.}
\label{fig:framework}
\end{figure*}

\section{Methodology}
In this section, we first illustrate our research problem formally and present task-related notations. 
% For convenient query, all symbols and brief descriptions used in this paper are offered in Table~\ref{tab:notation}. 
Along the processing line, we would introduce our Fine-grained Graph Propagation (FGP) framework in detail. 
%The overall framework of our proposed model is shown in 
Figure~\ref{fig:framework} illustrates the overall framework of FGP.
\subsection{Task-Related Notations}
We focus on the classic zero-shot learning problem. We use $\mathcal{C}, \mathcal{C}_{tr}, \mathcal{C}_{te}$ to denote the set of all classes, training classes, and test classes, respectively. Note that, here $\mathcal{C}_{tr} \cap \mathcal{C}_{te} = \emptyset$. Then we define $\mathcal{D}_{tr} = \{(\mathbf{X}_i, c_i) | i=1,2,...,|\mathcal{C}_{tr}| \}$ as the training image dataset, where $\mathbf{X}_i$ indicates the $i-$th training image and $c_i \in \mathcal{C}_{tr}$ the corresponding class label. Similarly, we get all image set $\mathcal{D}$ and the test image set $\mathcal{D}_{te}$. All classes in $\mathcal{C}$ are connected as nodes in a knowledge graph $G$, where each node $i$ is represented by a category name,  corresponding text description $T_i$, and images belong to this class. The symmetric adjacency matrix $A \in \mathbb{R}^{\mathcal{C} \times \mathcal{C}}$ denotes the connections between the classes in $G$, where $A_{i,j} = 1$ means two nodes $i$ and $j$ connected and vice versa. In this setting, zero-shot recognition aims to predict the class labels of a set of test data points to the set of classes $\mathcal{C}_{te}$. Unlike traditional classification, the test data points have to be assigned to previously unseen classes.

\subsection{Semantic Embeddings Generation}
Considering the human cognitive process, a person only needs to know two prior conditions if he intends to recognize the unseen object zebra: first, he has seen the horse or related horse photos as well as black and white stripe objects; second, this person has been told the zebra is a horse-like animal with black and white stripes. Here the semantic information of black and white stripes is the key point for humans or machines to identify zebra from various horses, which is also a typical fine-grained feature in this task. Then, how understanding the text description "black and white stripes" is one of the major tasks to simulate the human cognitive process for ZSL. To achieve this goal, in the first step, we design a Semantic Embeddings Generation (SEG) module to extract key semantic representations from text descriptions. It should be noted that there is a pretty necessary requirement to match the key semantics (such as stripe characteristic) with corresponding certain parts in the visual space (images, videos, or reality), which is the so-called (fine-grained) visual-semantic features alignment. In many specific datasets, the semantic alignment has been annotated with well-designed boolean attributes~\cite{liu2020attribute} (each sub-attribute is supposed to pay attention to a distinctive part). However, in the real-world dataset, the annotation for large-scale objects is unaffordable and not realistic. In order to achieve the alignment of the visual-semantic features in large-scale real-world datasets, we propose an unsupervised Semantic Embeddings Generation (SEG) module to extract the key semantic representations from text descriptions in the first step, which would benefit the following fine-grained visual-semantic alignment.

In the SEG module, we propose to utilize the unsupervised clustering algorithm to deal with the unlabeled text corpus that comes from the multimodal KG. As mentioned before, given the description $T_i$ of each concept $i \in \mathcal{C}$, we design and adopt several necessary preprocessing steps on the original text descriptions, including:
\begin{enumerate}[Step 1.]
\item Tokenization, lemmatization, converting uppercase to lowercase, and deleting stop words.
\item Using the dependency parsing tree in spacy tool$\footnote{https://spacy.io}$ to filter short phrases (nouns with modifiers, or modifiers, eg., adj + noun, noun, adj, etc) from sentences, such as ``white stripe''.
\item Using the pre-trained model Glove~\cite{pennington2014glove} to transfer all phrases into embedding vectors, where the embedding $\mathbf{p}_j$ of a phrase is produced by the mean pooling of all word embeddings that belong to it.
\end{enumerate}

For a fair comparison, we obtain word embeddings by the Glove model the same as DGP's approach~\cite{kampffmeyer2019rethinking} rather than other more advanced text models. After the preprocessing, we utilize the Kmeans clustering algorithm~\cite{macqueen1967some} to cluster all phrase embeddings into $K$ semantic groups. Along this line, we capture the centroids set: 
\begin{align}
\label{eq:v_c}
\Phi_v = \{\mathbf{v}^{c,k} \in \mathbb{R}^{d_c} | k=1,2,...,K\},
\end{align}
where these centroid $\mathbf{v}^{c,k}$ of obtained $K$ clusters are regarded as our key semantic embeddings, which are expected to have the capacity to concentrate on different and representative fine-grained attributes in the semantic space. From the view of machine understanding, the semantics of these key clusters may be abstract and do not necessarily correspond to the feature attributes concerned by human experts. We will explore the performance influence of the number of clusters in our experimental section.

\subsection{Fine-Grained Visual Knowledge Learning}
In this section, with the availability of original visual inputs and generated key semantic embeddings, we implement a dense attention module to learn the fine-grained visual knowledge, which ensures the alignment of visual-semantic space at the same time. 

The visual attention mechanism is able to generate a feature from the most relevant region of an image to match the key embedding and has been shown to be effective for image classification. Follow the setting of~\cite{huynh2020fine}, each image in $\mathbf{X}$ is divided into $R$ equal-size grid cells (regions) which are denoted by $\{I^r | r\in R\}$. Here we use a ResNet-50~\cite{he2016deep} pretrained on ImageNet~\cite{deng2009imagenet} to catch the feature vector $\mathbf{f}^r \in \mathbb{R}^{d_f}=f(I_r)$ of region $r$, where $f(\cdot)$ denotes the ResNet-50 network. For a sample, we define the query value of the attention mechanism by $\mathbf{v}^c$ and key values by $\{ \mathbf{f}^r | r \in R \}$, then the weight of each query on keys can be calculated as Eq.~(\ref{eq:alpha}),
\begin{align}
\label{eq:alpha}
\alpha_k^r = \frac{\mathrm{exp}(\mathbf{v}^{c,k} \mathbf{w}_{\alpha} \mathbf{f}^r)}{ \mathrm{exp}(\sum_{r^{\prime} = 1}^R \mathbf{v}^{c,k} \mathbf{w}_{\alpha} \mathbf{f}^{r^{\prime}}) },
\end{align}
where $\mathbf{w}_{\alpha} \in \mathbb{R}^{d_c \times d_f}$ is the parameter to expand the dimension of $\mathbf{v}^{c,k}$ to be equal to the dimension of visual feature $\mathbf{f}^r$. In the weight distribution $\alpha_k = \{ \alpha_k^1, \alpha_k^2,...,\alpha_k^R \}$ of key semantic embedding $\mathbf{v}^{c,k}$, $\alpha_k^r \in [ 0, 1 ]$ and $\sum_{r=1}^{R}\alpha_k^r = 1$ means that our model can select different regions according to diverse weight values. Hence we carry out the selection procedure easily and get the output as follows:
\begin{align}
\label{eq:select}
\mathbf{e_k} = \sum_{r=1}^R \alpha_k^r \mathbf{f}^r .
\end{align}

Now we obtain the attention-based representation. Each state $\mathbf{e}_k$ contains the enhanced certain region which focuses on different topics. That is to say, our method possesses the ability of extracting fine-grained features from rough information. We aim to further ensure that the key semantic embedding recognizes the required key visual object which corresponds to the description of text data. To achieve this goal, we design a self-calibration loss that restrains the distance between the fine-grained feature and corresponding semantic embedding. In detail, we assume that the representations' correlation of focused regions from different images that belong to the same key semantic embedding should be relatively close compared with other matched pair types. For example, if the semantic key concentrates on the ``\textbf{yellow body}'' related topic, the distance to this key embedding from text embedding of class \textbf{Lion} and \textbf{Orange Cat} should be closer than that from \textbf{Bear} or \textbf{Panda}, and so on. Along this line, for a class $c$, we first calculate the cosine distances $dis_c \in \mathbb{R}^{K}$ between its text embedding and our key semantic embeddings. Regarding $dis_c$ as a new label, we just need to approach the similarity of each fine-grained feature $\{ \mathbf{e}_k | k \in K \}$ (for convenience, here we omit the sample id $i$ of $\mathbf{e}_k$ obtained from the sample with label $c$) and $dis_c$ constantly for each class $c$.
\begin{align}
\label{eq:self-calibration}
\mathcal{L}_{sc} = \frac{1}{2K} \sum_{k=1}^K \parallel \mathrm{ReLU}(\mathbf{e}_k \mathbf{w}_{sc}) - {dis_c}^k \parallel^2.
\end{align}
The self-calibration loss $\mathcal{L}_{sc}$ is shown in Eq. (\ref{eq:self-calibration}), where $\mathbf{w}_{sc} \in \mathbb{R}^{d_f \times 1}$ denotes the training mapping parameter. Actually, with the optimization iterations, the similar fine-grained regions in various images would be aligned to the correlative key semantic embedding step by step. In this way, our model gradually understands what kind of visual feature is ``yellow body''. Having got this far, more than half of the motivation for fine-grained information alignment has been achieved. Note that the optimization of all parameters in this section would be executed in a fine-tuning process.

We will freeze the parameters of ResNet-50 and fine-tune our model on the ImageNet-2012 dataset. Specifically, for $\{ \mathbf{e}_k | k \in K \}$, we implement individual fully-connected layer ${\mathbf{w}_f^k}^{\top} \in \mathbb{R}^{d_f \times |\mathcal{C}_{tr}|}$ as the classifier for each of them. So the final output is: 
\begin{align}
\label{eq:out}
\mathbf{o} = \mathbf{e}_g {\mathbf{w}_g}^{\top} + \mathbf{b_g} + \sum_{k=1}^K \mathbf{e}_k {\mathbf{w}_f^k}^{\top} + \mathbf{b}_k ,
\end{align}
where $\mathbf{e}_g$ and $\mathbf{w}_g$ denote the global feature extracted by ResNet-50 and corresponding global classifier, respectively. The final cross-entropy loss of our zero-shot classification task is defined as follows:
\begin{align}
\label{eq:entropyloss}
\mathcal{L}_{ce} = - \frac{1}{|\mathcal{D}_{tr}|} \sum_{i \in \mathcal{D}_{tr}} \mathrm{log} \frac{\mathrm{exp}(\mathbf{o}_i^{c_i})}{\sum_{c^{\prime} \in \mathcal{C}_{tr}} \mathrm{exp}( \mathbf{o}_i^{c^{\prime}})},
\end{align}
where $c_i$ denotes the true class label of sample $i$. In the fine-tuning stage, we optimize the loss $\mathcal{L}_{sc}$ for all samples and $\mathcal{L}_{ce}$ simultaneously until they converge to stable status.

% \renewcommand{\algorithmicrequire}{\textbf{Input:}}
% \renewcommand{\algorithmicensure}{\textbf{Output:}}
% \begin{algorithm}[t]
% \caption{algorithm caption}
% \Require input parameters A, B, C\\
% \Ensure output result
% \begin{algorithmic}[1]
% \State some description
% \State \Return result
% \end{algorithmic}
% \end{algorithm}
\subsection{Structural Knowledge Learning}
To learn the relationships between unseen and seen classes, we draw support from the structural association in the knowledge graph by GCN as well as~\cite{kampffmeyer2019rethinking}. It is a semi-supervised message propagation approach that aggregates implicit and explicit correlations of nodes (classes) into the node representations. In detail, we input the complete text embedding of each class as the initial state of GCN to predict the last layer weight (classifier) of CNN (e.g., ResNet) and then replace the original weight with obtained prediction vector. Because the GCN network could also learn the projection weights for unseen classes according to its semi-supervised training style, the classification framework obtains the ability to recognize unseen concepts after a fine-tuning of the CNN's parameters (the new classifiers learned by GCN are frozen in the fine-tuning process). 

\begin{figure}[t]
  \centering
  \subfigure[Previous method]{
    % \label{fig:kmeans1} %% label for first subfigure
    \includegraphics[width=0.195\textwidth]{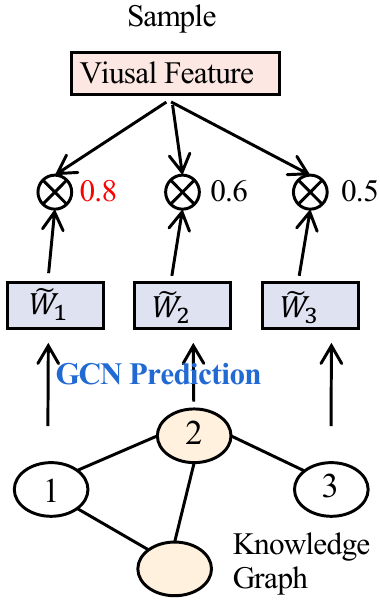}}
  \subfigure[Our fine-grained method]{
    % \label{fig:kmeans2} %% label for second subfigure
    \includegraphics[width=0.263\textwidth]{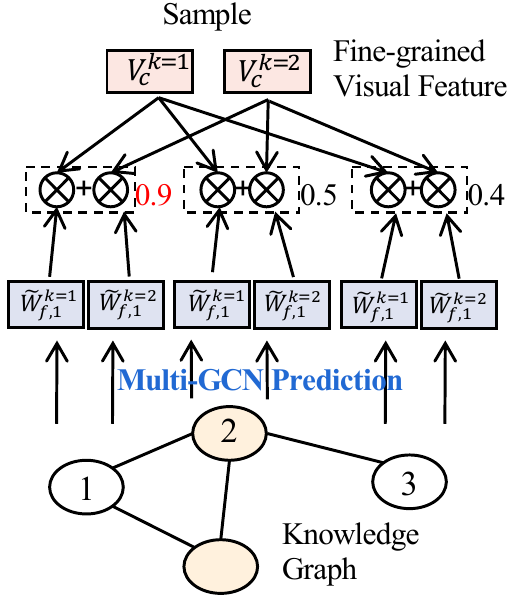}}
  \caption{The difference between previous rough GCN prediction and our fine-grained aligned prediction. On the knowledge graph, the orange nodes denote seen classes with labels and the white nodes denote unseen classes.}
  \label{fig:GCNmodel} %% label for entire figure
\end{figure}
Different from previous KG-based models, in the weight fitting step, we have to maintain different classifiers for several fine-grained features with different focused topics. It is obviously inappropriate to use a single GCN with the same state to predict the weight parameters of all classifiers. Meanwhile, we still cannot throw away the visual-semantic alignment condition at this stage. To this end, we propose an effective but not complicated Multi-GCN network on the knowledge graph to fit our target parameters. We define each unique GCN of Multi-GCN as:
\begin{align}
\label{eq:gcn}
\mathbf{H}^{l+1}_{k} = \sigma (D^{-1}A\mathbf{H}^{l}_k\mathbf{\theta}^{l}_k), k=1,2,...,K,g,
\end{align}
where the symmetric adjacency matrix $A \in \mathbb{R}^{\mathcal{C} \times \mathcal{C}}$ denotes the connections between the classes in the knowledge graph, which also includes self-loops. $D_{i,i}=\sum_{j}A_{i,j}$ is a degree matrix that normalizes rows in $A$ to ensure that the value scale of feature representations is not modified by $A$. $\mathbf{\theta}_k^{l} \in \mathbb{R}^{d_c \times F}$ denotes the trainable weight matrix for layer $l$ with $F$ corresponding to the number of learned filters. $\sigma$ indicates the nonlinear activation function Leaky ReLU. Besides, $\mathbf{H}^{l}_k$ represents the $l$-th layer's hidden state for classifiers in the $k$-th fine-grained channel. Here for each node $i$, the initial node embedding is defined as follows:
\begin{gather}
\label{eq:p_k}
\mathbf{H}^0_{k,i} = \mathrm{MeanPooling}(\mathbf{P}_{k,i}),\\ \mathbf{P}_{k, i} = \{ \mathbf{p}_j | dis_c(\mathbf{p}_j, \mathbf{v}^{c,k}) \leq \forall dis_c(\mathbf{p}_j, \mathbf{v}^{c,\hat{k} \neq k} ), \nonumber \\  \hat{k} \leq K, \mathbf{p}_j \in T_i \},
\end{gather}
where $dis_c(\cdot)$ is also the cosine distance to calculate vectors' similarity, and $\mathbf{P}_{k,i}$ denotes all Glove-based phrase embeddings in the text data $T_i$ that more close to the key semantic embedding $\mathbf{v}^{c,k}$ than others ($\{ \mathbf{v}^{c,\hat{k}} | \hat{k} \neq k \}$). In the condition mentioned above, the prediction flow based on the knowledge graph is aligned with the fine-grained level knowledge, which further avoids the information interference of mistaken visual-text (semantic) matchings.

In the loss function $\mathcal{L}_{gcn}$, the Multi-GCN model is trained to predict the classifier weights for the seen classes by optimizing $\mathcal{L}_{gcn}$, where each $\mathbf{\widetilde{w}} \in \mathbb{R}^{\mathcal{C}_{tr} \times d_f}$ is the prediction of the GCN for the known classes and corresponds to the $\mathcal{C}_{tr}$ rows of the GCN output.
\begin{equation}
\begin{aligned}
\label{eq:gcnloss}
\mathcal{L}_{gcn} = \frac{1}{2\mathcal{C}_{tr}(K+1)} [ \sum_{i=1}^{\mathcal{C}_{tr}} \sum_{j=1}^{d_f}(\mathbf{w}_{g,i,j} - \mathbf{\widetilde{w}}_{g,i,j}) + \\ \sum_{k=1}^K  \sum_{i=1}^{\mathcal{C}_{tr}} \sum_{j=1}^{d_f}(\mathbf{w}_{f,i,j}^k - \mathbf{\widetilde{w}}_{f,i,j}^k) ].
\end{aligned}
\end{equation}

To illustrate the GCN-based prediction more clearly, we give the structure comparison in Figure~\ref{fig:GCNmodel}, where each number in the node of KG denotes a class id and the red scores denote true predictions. According to this figure, we can find that the classifiers of unseen classes are also able to be learned within the semi-supervised learning process, which makes the zero-shot recognition for unseen classes possible. And our proposed model achieves the individual fine-grained prediction with sub-text-visual alignment to promote the performance. In other words, for each enhanced fine-grained visual feature with a certain semantic topic, we set an individual channel for GCN to implement semi-supervised training.

Finally, for more clear illustration, the whole algorithm flow of our multimodal framework is presented in \textbf{Algorithm}~\ref{algorithm1}, where step 3$\sim$4 denotes the Semantic Embeddings Generation approach, step 5 describes the training way of Fine-grained Visual Knowledge Learning, and step 6$\sim$8 introduces the Structural Knowledge Learning method and the final classification procedure.

\begin{table*}[t]
\centering
\def\arraystretch{.92}
\caption{Top-k accuracy results for ZSL setting(only testing on unseen classes) on three subsets based on the ImageNet dataset.}
\label{tab:results}
\begin{tabular}{cc|ccccc|ccccc|ccccc}
\toprule
\multirow{2}{*}{ Method type } & \multirow{2}{*}{ Model } & \multicolumn{5}{c}{ Hit@k (\%) on 2-hops } & \multicolumn{5}{c}{ Hit@k (\%) on 3-hops} & \multicolumn{5}{c}{ Hit@k (\%) on All}\\
& & 1 & 2 & 5 & 10 & 20 & 1 & 2 & 5 & 10 & 20 & 1 & 2 & 5 & 10 & 20 \\
\midrule 
\multirow{5}{*}{ \shortstack{Embedding-\\based method}} & DeViSE & $6.7$ & $11.2$ & $19.4$ & $28.1$ & $38.3$ & 2.1 & 3.5 & 6.3 & 9.5 & 14.1 & 1.0 & 1.8 & 3.0 & 4.6 & 7.1\\
& ConSE & $8.3$ & $12.9$ & $21.8$ & $30.9$ & $41.7$ & 2.6 & 4.1 & 7.3 & 11.1& 16.4 & 1.3 & 2.1 & 3.8 & 5.8 & 8.7 \\
& SYNC & $10.5$ & $17.7$ & $28.6$ & $40.1$ & $52.0$ & 2.9 & 4.9 & 9.2 & 14.2 & 20.9 & 1.4 & 2.4 & 4.5 & 7.1 & 10.9\\
& EXEM & $12.5$ & $19.5$ & $32.3$ & $43.7$ & $55.2$ & 3.6 & 5.9 & 10.7 & 16.1 & 23.1 & 1.8 & 2.9 & 5.3 & 8.2 & 12.2\\
\midrule
\multirow{6}{*}{\shortstack{Knowledge \\Graph-based\\ method}} & GCNZ & $19.8$ & $33.3$ & $53.2$ & $65.4$ & $74.6$ & 4.1 & 7.5 & 14.2 & 20.2 & 27.7 & 1.8 & 3.3 & 6.3 &  9.1 & 12.7\\
& DGP & $26.6$ & $40.7$ & $60.3$ & $72.3$ & $81.3$ & 6.3 & 10.7 & 19.3 & 27.7 & 37.7 & 3.0 & 5.0 & 9.3 & 13.9 & 19.8\\
& GPCL & $\mathbf{2 8 . 4}$ & $\mathbf{4 3 . 0}$ & $\mathbf{6 2 . 6}$ & $\mathbf{7 4 . 5}$ & $\mathbf{8 2 . 9}$ & 7.0 & \textbf{11.7} & 20.7 & 29.2 & 39.0 & 3.3 & 5.6 & 10.1 & \textbf{14.7} & 20.5\\
& HVEL & $13.3$ & $20.8$ & $39.2$ & $52.7$ & $62.4$ & 6.5 & 10.6 & 18.8 & 25.8 & 35.2 & \textbf{3.7} & \textbf{5.9} & 10.3 & 13.0 & 16.4\\
& MGFN & 26.9 & 40.9 & 60.3 & 72.7 & 81.4 & 6.3 & 10.8 & 19.6 & 27.9 & 37.9  & 3.2 & 5.7 & 9.6 & 14.1 & 19.9\\
\cmidrule(l){2-17}
& FGP (ours) & $26.4$ & $41.2$ & $61.0$ & $72.8$ & $81.7$ & \textbf{7.3} & 11.0 & \textbf{21.3} & \textbf{29.8} & \textbf{39.4} & \textbf{3.7} & 5.5 & \textbf{10.5} & \textbf{14.7} & \textbf{20.9}\\
\bottomrule
\end{tabular}
\end{table*}

\begin{algorithm}[t]
  \SetAlgoLined
  \SetKwInOut{Input}{input}\SetKwInOut{Output}{output}
  \Input{The set of all classes $\mathcal{C}$, training classes $\mathcal{C}_{tr}$, and test classes $\mathcal{C}_{te}$. The training image dataset $\mathcal{D}_{tr}$, test image set $\mathcal{D}_{te}$.
	The knowledge graph $G$, the text description $T_i$ of each class $i$}
  \Output{Prediction values for test dataset}
  Get pre-trained ResNet-50 backbone $\Theta_1$\;
  Randomly initialize attention parameter $\Theta_2$, classification layer $\{\mathbf{w}_f, \mathbf{w}_g\}$, Multi-GCN parameter $\Theta_3$\;
  Preprocess each original text $T_i$ into phrase embeddings\;
  Generate $K$ semantic clusters and centroids $\mathbf{v}^{c}$ by Kmeans\;
  Freeze ResNet-50 parameter $\Theta_1$ and fine-tune parameters $\{ \Theta_2, \mathbf{w}_f, \mathbf{w}_g\}$ with query $\mathbf{v}^{c}$ on $\mathcal{D}_{tr}$ in the fine-grained visual-semantic alignment process\;
  Fit classifier weights $\{\mathbf{w}_f, \mathbf{w}_g\}$ by the output matrix $\{\mathbf{\widetilde{w}}_f, \mathbf{\widetilde{w}}_g\}$ of Multi-GCN $\Theta_3$ trained with phrase embedding inputs of $\mathcal{C}_{tr}$ on $G$\;
  Replace $\{\mathbf{w}_f, \mathbf{w}_g\}$ with $\{\mathbf{\widetilde{w}}_f, \mathbf{\widetilde{w}}_g\}$\;
  Fine-tune $\Theta_1$ on $\mathcal{D}_{tr}$, then output classification values $\mathbf{\widetilde{y}}_{te}$ for all samples in $\mathcal{D}_{te}$\;
  \textbf{return} $\mathbf{\widetilde{y}}_{te}$;
  \caption{Fine-grained Graph Propagation}
   \label{algorithm1}  
\end{algorithm}

\section{Experiments}
% In the experimental section, we first introduce two public datasets. Then several competitive baselines would be described in detail. The experimental results of several evaluations are subsequently reported, which consist of performance comparison, ablation studies, and hyperparameter sensitivity testing. Finally, we design an attention-value visualization case to check the alignment mechanism of fine-grained features.
\subsection{Datasets}
\subsubsection{ImageNet} We conduct the experiments of our model and baselines on the ImageNet dataset~\footnote{https://image-net.org/download.php}. Following the setting of \cite{frome2013devise,kampffmeyer2019rethinking}, we adopt the train/test split of them and evaluate the zero-shot classification on the large 21K ImageNet dataset. Since the fine-tuning process is executed on the ImageNet 2012 1K classes, the test datasets consist of three parts: ``2-hops'', ``3-hops'', and ``All''. Hops refer to the distance that classes are away from the ImageNet 2012 1K classes in the ImageNet hierarchy and thus is a measure of how far unseen classes are away from seen classes. ``2-hops'' includes roughly 1.5K classes within two hops from the seen classes, while ``3-hops'' contains 7.8K classes. ``All'' contains close to 21K classes. All these classes are not contained in ImageNet 2012, which was used to pre-train the ResNet-50 model. What's more, we further evaluate the performance when training categories are included as potential labels. Here we called the splits as ``2-hops+1K'', ``3-hops+1K'', ``All+1K''. We will evaluate the Top 1$\sim$20 accuracy on these three (sub) datasets.
\subsubsection{AWA2} We follow the setting of \cite{kampffmeyer2019rethinking} to test the AWA2 dataset~\footnote{http://cvml.ist.ac.at/AwA2/}. It consists of 50 animal classes, with a total of 37,322 images and an average of 746 per class. The split ensures that there is no overlap between the test classes and the ImageNet 2012 dataset, where 40 classes are used for training and 10 for testing. AWA2 test classes are contained in the 21K ImageNet and several of the training classes (24 out of 40) that are in the proposed split overlap with the ImageNet 2012 dataset. Note that, same as the setting of ~\cite{kampffmeyer2019rethinking}, we also only test the Top-1 accuracy of each model on the zero-shot setting since the relatively small scale of this dataset and the number of unseen classes is not enough to try large indicators.

\begin{table*}[t]
\centering
\def\arraystretch{.95}
\caption{Top-k accuracy results with General ZSL setting(testing instances includes seen and unseen classes simultaneously) on three subsets based on the ImageNet dataset. $^\ddagger$ denotes the result from \cite{wang2018zero}. Here we also follow~\cite{kampffmeyer2019rethinking,liu2020hyperbolic} do not use SYNC and EXEM due to the algorithm characteristics. }
\label{tab:results_generalized}
\begin{tabular}{cc|ccccc|ccccc|ccccc}
\toprule
\multirow{2}{*}{ Method type } & \multirow{2}{*}{ Model } & \multicolumn{5}{c}{ Hit@k (\%) on 2-hops } & \multicolumn{5}{c}{ Hit@k (\%) on 3-hops} & \multicolumn{5}{c}{ Hit@k (\%) on All}\\
& & 1 & 2 & 5 & 10 & 20 & 1 & 2 & 5 & 10 & 20 & 1 & 2 & 5 & 10 & 20 \\
\midrule
\multirow{3}{*}{ \shortstack{Embedding-\\based method}} & DeViSE & 0.8 & 2.7 & 7.9 & 14.2 & 22.7 & 0.5 & 1.4 & 3.4 & 5.9 & 9.7 & 0.3 & 0.8 & 1.9 & 3.2 & 5.3\\
& ConSE & 0.3 & 6.2 & 17.0 & 24.9 & 33.5 & 0.2 & 2.2 & 5.9 & 9.7 & 14.3 & 0.2 & 1.2 & 3.0 & 5.0 & 7.5 \\
& ConSE$^\ddagger$ & 0.1 & 11.2 & 24.3 & 29.1 & 32.7 & 0.2 & 3.2 & 7.3 & 10.0 & 12.2 & 0.1 & 1.5 & 3.5 & 4.9 & 6.2\\
\midrule
\multirow{6}{*}{\shortstack{Knowledge \\Graph-based\\ method}} & GCNZ & 9.7 & 20.4 & 42.6 & 57.0 & 68.2 & 2.2 & 5.1 & 11.9 & 18.0 & 25.6 & 1.0 & 2.3 & 5.3 & 8.1 & 11.7\\
& DGP & 10.3 & 26.4 & 50.3 & 65.2 & 76.0 & 2.9 & 7.1 & 16.1 & 24.9 & 35.1 & 1.4 & 3.4 & 7.9 & 12.6 & 18.7\\
& GPCL & 7.0 & 26.8 & 52.5 & \textbf{67.5} & \textbf{77.9}  & 2.0 & 7.1 & 17.3 & 26.2 & \textbf{36.5} & 1.0 & 3.4 & 8.5 & 13.2 & \textbf{19.3}\\
& HVEL & 6.4 & 11.9 & 27.2 & 35.3 & 45.2 & 3.6 & \textbf{8.7} & 15.3 & 20.5 & 29.1 & 2.2 & 4.6 & 9.2 & 12.7 & 15.5\\
& MGFN & \textbf{13.5} & 28.1 & 51.1 & 65.4 & 76.1 & 3.7 & 7.3 & 16.5 & 25.3 & 35.9 & 1.8 & 3.5 & 8.2 & 12.8 & 18.9\\
\cmidrule(l){2-17}
& FGP (ours) & 12.0 & \textbf{28.8} & \textbf{53.0} & 65.4 & 75.8 & \textbf{4.0} & \textbf{8.7} & \textbf{17.9} & \textbf{26.6} & 36.2 & \textbf{2.5} & \textbf{4.8} &  \textbf{9.4} & \textbf{13.5} & 18.4\\
\bottomrule
\end{tabular}
\end{table*}

\subsection{Baselines}
We compare our model to some representative approaches which are suitable for non-attribute datasets. Specifically, $\textbf{DeViSE}$~\cite{frome2013devise} leverages textual data to learn semantic relationships between different labels, and explicitly maps images into a rich semantic embedding space. $\textbf{ConSE}$~\cite{norouzi2013zero} projects visual features into the semantic space as a convex combination of several semantic embeddings of the $T$ nearest seen classes which are weighted by the probabilities that the image belongs to seen classes. $\textbf{EXEM}$~\cite{changpinyo2017predicting} regards the cluster centers of visual feature vectors as the target semantic representations and leverages structural relations well on the cluster. $\textbf{SYNC}$~\cite{changpinyo2016synthesized} aims to align the semantic space with the visual space via phantom object classes to connect seen and unseen classes. $\textbf{GCNZ}$~\cite{wang2018zero} uses both semantic embeddings and categorical relationships to predict the final classifiers. GCNZ inputs semantic embedding for each concept node (representing visual category) in the knowledge graph. $\textbf{DGP}$~\cite{kampffmeyer2019rethinking} is a KG-based graph embedding model which explicitly exploits the hierarchical structure of the KG to perform ZSL by propagating knowledge through the dense connectivity structure. 
$\textbf{GPCL}$~\cite{wang2021zero} exploits multiple relationships among different categories of KG for zero-shot learning by employing graph convolutional representation and contrastive learning techniques.
$\textbf{HVEL}$~\cite{liu2020hyperbolic} further learns hierarchical-aware image embeddings in hyperbolic space for zero-shot learning, which is capable of preserving the hierarchical structure of semantic classes. $\textbf{MGFN}$~\cite{wei2022semantic} develops a multi-granularity fusion network that integrates discriminative information from multiple GCN branches.

\begin{figure}[t]
\centering
\includegraphics[width=0.48\textwidth]{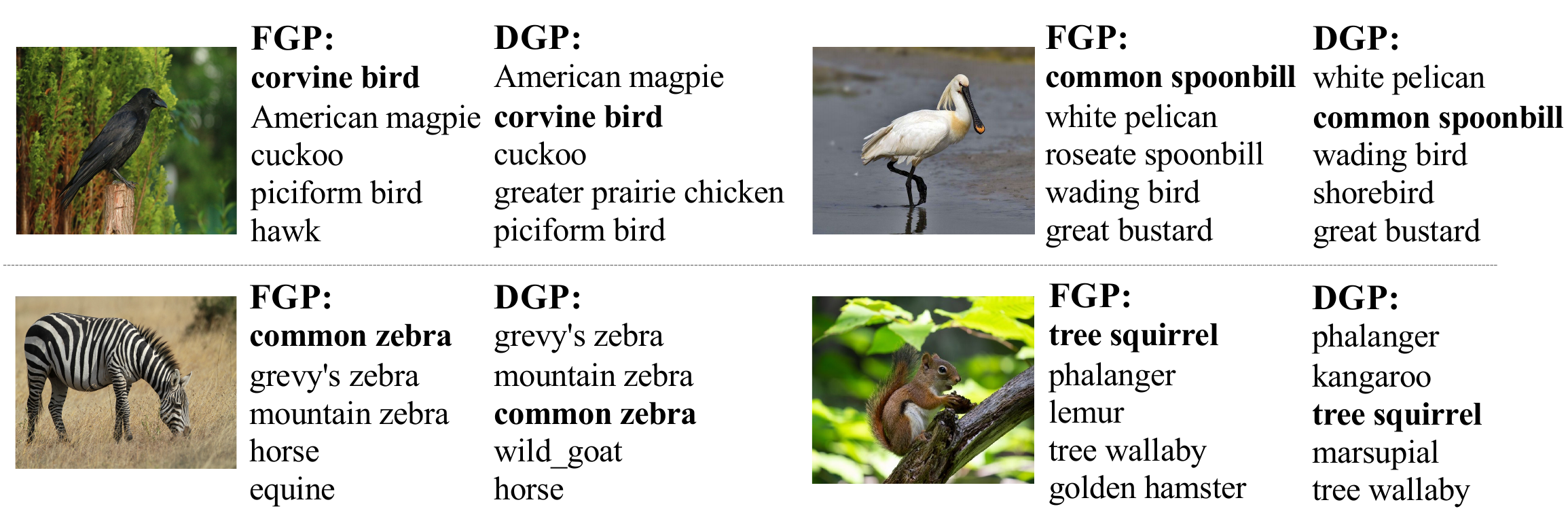}
\caption{The qualitative comparison cases of top-5 classification results  on some unseen categories under the ZSL setting. The correct category is marked as bold in each case. }
\label{fig:sample}
% \vspace{-0.3cm}
\end{figure}

\begin{table}[t]
\centering
\setlength{\tabcolsep}{10pt}
\def\arraystretch{.7}
\caption{Top-1 accuracy (\%) results for zero-shot recognition on the AWA2 dataset.}
\label{tab:results_awa}
\begin{tabular}{cc|c|c}%c}
%\cmidrule[1.5pt]{1-3}
\toprule
\bf Method type & \bf Model &  \bf ZSL & \bf GZSL \\
\midrule
\multirow{4}{*}{ \shortstack{Embedding-\\based method}} & ConSE & 44.5 & 5.8\\
& DeViSE & 59.7 & 16.0\\
& SYNC & 46.6 & -\\
& EXEM & 55.6 & -\\
\midrule
\multirow{6}{*}{ \shortstack{Knowledge \\Graph-based\\ method}}
& GCNZ & 70.7 & 37.3\\
& DGP & 77.3 &  40.2\\
& GPCL & 77.8 & 41.4\\
& HVEL & 77.0 & 39.8\\
& MGFN & 74.2 & 38.3\\
\cmidrule(l){2-4}
& FGP (ours) & {\bf 79.1} & {\bf 43.3}\\
\bottomrule
\end{tabular}
\end{table}

\subsection{Implementation Details}
We use a ResNet-50 model that has been pre-trained on the ImageNet 2012 dataset. We extract a feature map at the last convolutional layer whose size is $7 \times 7 \times 2048$ and treat it as a
set of features from $R = 7 \times 7$ regions. The GloVe model trained on the Wikipedia dataset is utilized to obtain the feature representation of concepts' descriptions in the KG (we use the descriptions from WordNet~\cite{miller1998wordnet} in this paper). Every single GCN in the Multi-GCN model consists of layers with hidden dimensions of 2048 and the final output dimension corresponds to the number of weights in the last layer of the ResNet-50 architecture, 2049 for weights and bias. In addition, we regularize the outputs into similar ranges by L2-Normalization. Similarly, we also normalize the ground truth weights produced by CNN. To avoid over-fitting, we implement Dropout~\cite{srivastava2014dropout} with a dropout rate of 0.5 in each layer of GCN. Our Multi-GCN model is trained for 3000 epochs with a learning rate of 0.001 and weight decay of 0.0005 by the Adam optimizer~\cite{kingma2014adam}. The negative slope of leaky ReLU is 0.2. The proposed FGP model is performed on an NVIDIA Tesla V100 GPU. Fine-tuning is done for 20 epochs using SGD with a learning rate of 0.0001 and momentum of 0.9. What's more, we explore the influence of $K$, the number of key semantic embeddings, and $L$, the number of GCN layers in the ablation studies. The dimension $d_c=300$ of input vectors of Kmeans is the same as the output dimension of Glove.

\subsection{Performance Comparison}
\subsubsection{Quantitative Performance Comparison}
We report the quantitative results of ZSL testing and General ZSL testing on ImageNet in Table~\ref{tab:results} and Table~\ref{tab:results_generalized}, respectively. To give a more intuitive comparison, some qualitative comparison results are shown in the next subsection. The results of baselines are taken from~\cite{kampffmeyer2019rethinking,liu2020hyperbolic,wang2021zero}. Following the setting of them, we utilize the Top-k accuracy to evaluate the model as well, and the EXEM model and the SYNC model are only used on the ZSL task due to their characteristics. 
Overall, our proposed model FGP achieves obviously better results than other comparing methods on most of the testing tasks. Compared with previous powerful KG-based models DGP, GPCL, MGFN, and HVEL, with the help of fine-grained knowledge, our model reflects a more comprehensive performance on different scale settings. It is easy to find that HVEL shows a serious bias on the ``All'' test data, which outperforms all baselines including DGP on this experimental part but is significantly weaker than DGP, GCNZ on both ZSL and GZSL evaluations when recognizing the ``2-hops'' dataset. Similarly, the GPCL model performs well on the ``2-hops'' ZSL task but is not enough good as our model on other settings, especially the GZSL tasks. On the contrary, although the bias tendency of DGP is not so critical as the same as former, this model still performs not good on the biggest dataset ``All''. 

Going deeper into the results produced by FGP, it beats all rivals in the ``2-hops+1K'' and ``All + 1K'' tasks and reaches the highest Top-1 scores 4.0\% and 2.5\%, which provide 8.1\% and 13.6\% relative improvement over the previous state-of-the-art. Besides, on the ``3-hops'' and ``All'' datasets of the ZSL setting, FGP still shows the best performance. Specifically, FGP surpasses all baselines on most indicators with a distinct margin, where HVEL is slightly better than our model on Top-2 accuracy but not ideal on other indicators. To summarize, the single-modality KG model learns the semantic transfer by relationships, which gradually loses the recognition ability as the number of hops increases. And the fine-grained features captured by FGP (multimodal KG) alleviate this shortage.

On the other hand, the further exploration results on the AWA2 dataset are also shown in Table~\ref{tab:results_awa}. Here we follow the setting of DGP~\cite{kampffmeyer2019rethinking} to test the Top-1 accuracy of each model since the relatively small scale of this dataset and the number of unseen classes is not enough to try large indicators. Obviously, our proposed model gets the best result of 79.1\% and 43.3\% accuracy on the ZSL and GZSL settings, respectively.

\subsubsection{Qualitative Comparison}
To give a more intuitive comparison, some qualitative comparison results produced by models are shown in Figure~\ref{fig:sample}. The samples come from the ``2-hops'' set and their corresponding predicting tags are produced by DGP and FGP under the ZSL setting (only testing on unseen classes), since the DGP model inspires our motivation. Intuitively, from the top-5 classification results of each testing image, we observe that DGP and FGP generally provide coherent top-5 results due to the relationships of classes in the knowledge graph.  According to Figure~\ref{fig:sample}, with the help of more fine-grained visual features and textual features, our model obtains better performance in many samples on zero-shot classification. For example, although ``common zebra'' and ``grevy's zebra'' are very similar in body shape, they can easily be distinguished by the shape of their ears, where the former has pointed ears and the latter has round ears.

\begin{figure}[t]
  \centering
  \subfigure[ZSL on ``All'']{
    % \label{fig:ab_ZSL} %% label for first subfigure
    \includegraphics[width=0.45\columnwidth]{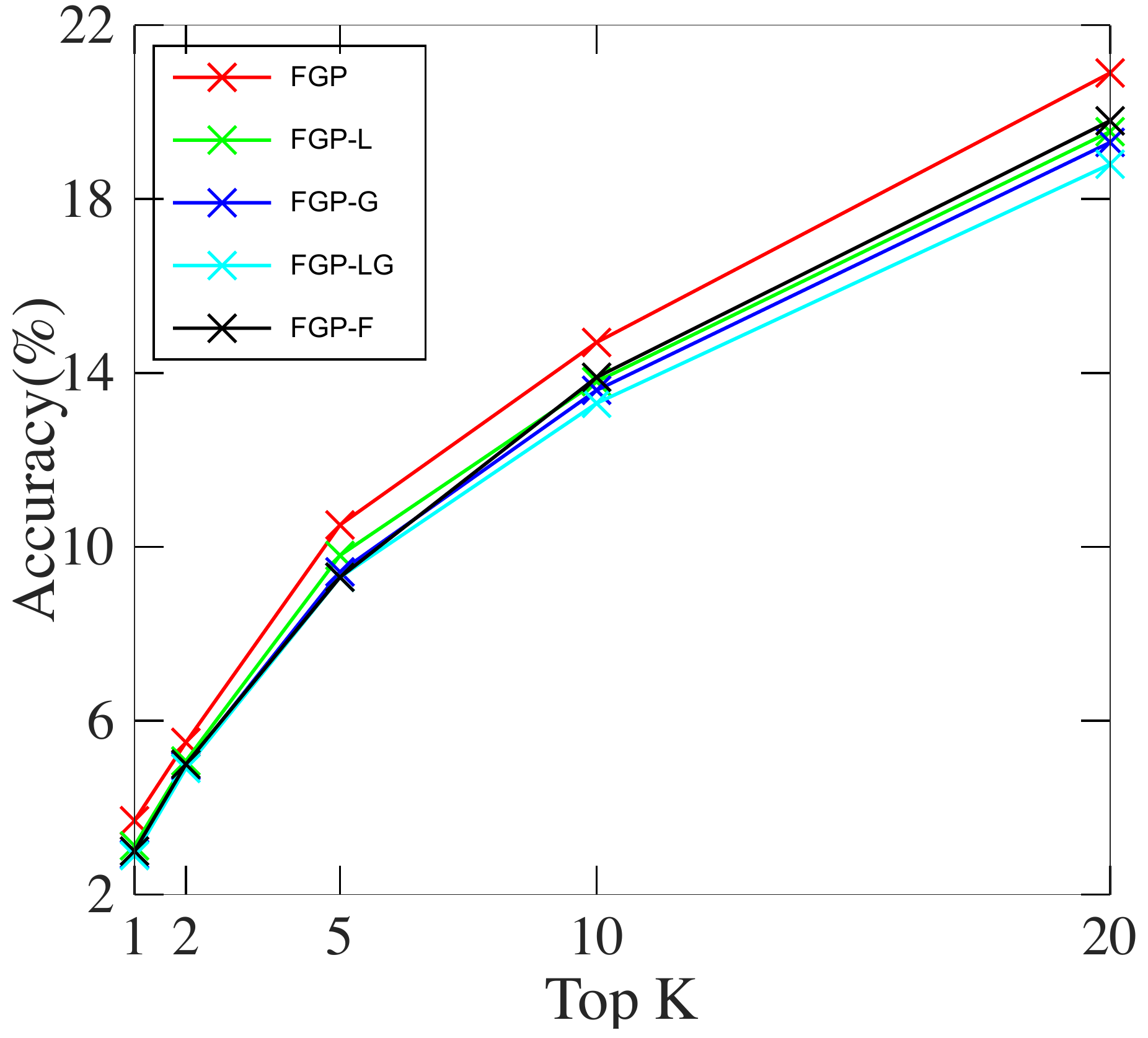}}
  \subfigure[GZSL on ``All'']{
    % \label{fig:ab_GZSL} %% label for second subfigure
    \includegraphics[width=0.45\columnwidth]{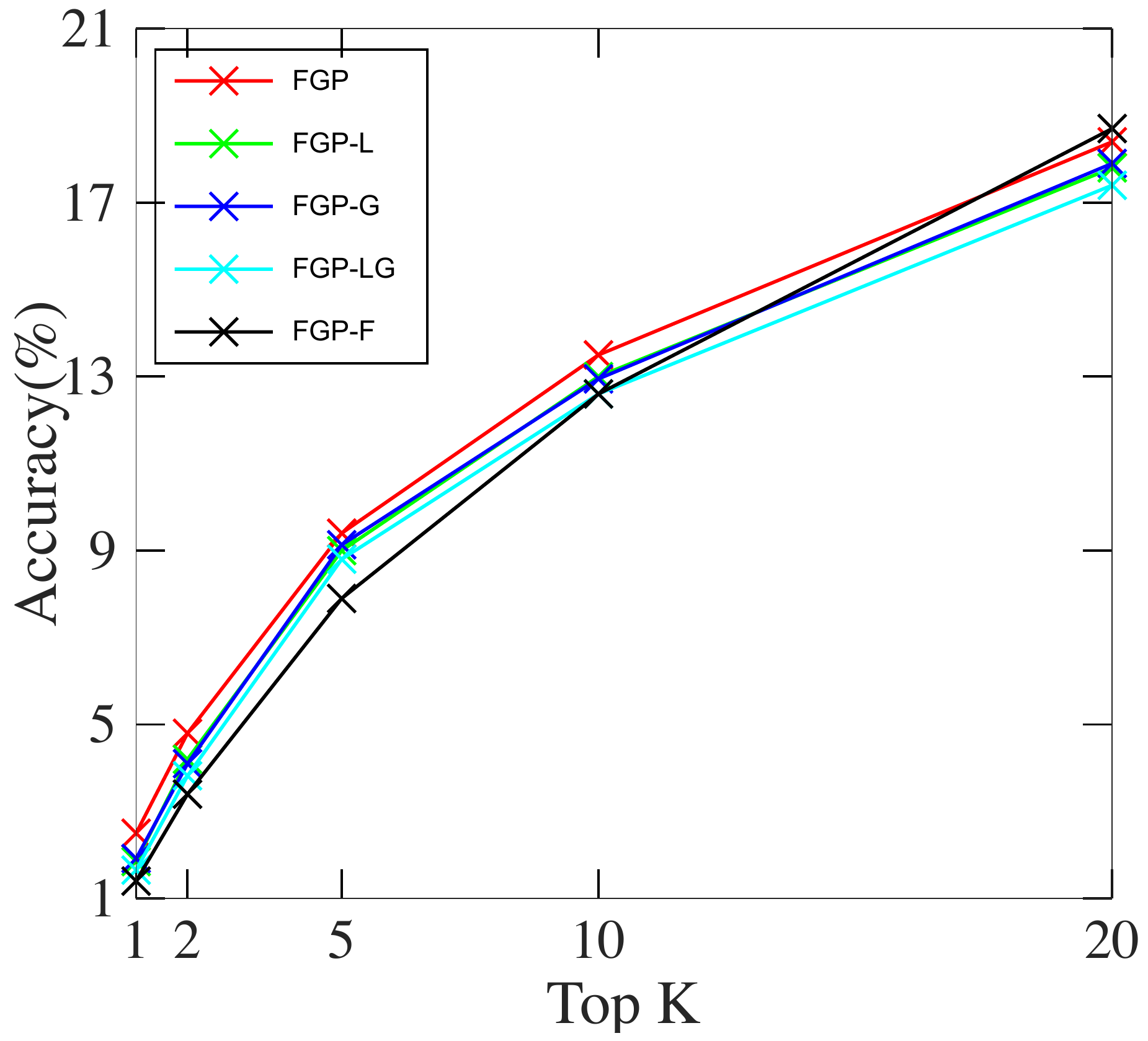}}
  \subfigure[ZSL on AWA2]{
    % \label{fig:ab_ZSL_awa} %% label for second subfigure
    \includegraphics[width=0.45\columnwidth]{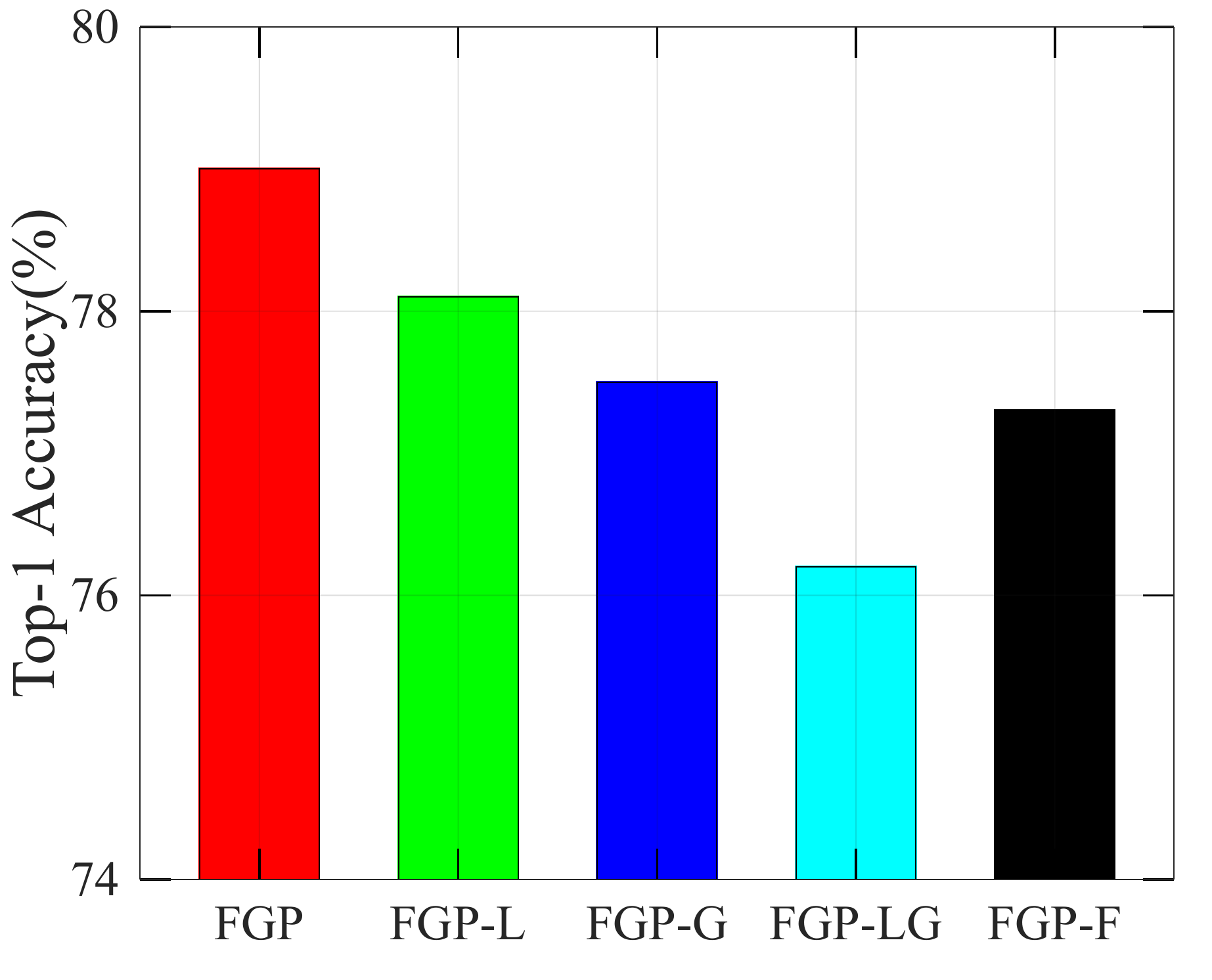}}
  \subfigure[GZSL on AWA2]{
    % \label{fig:ab_GZSL_awa} %% label for second subfigure
    \includegraphics[width=0.45\columnwidth]{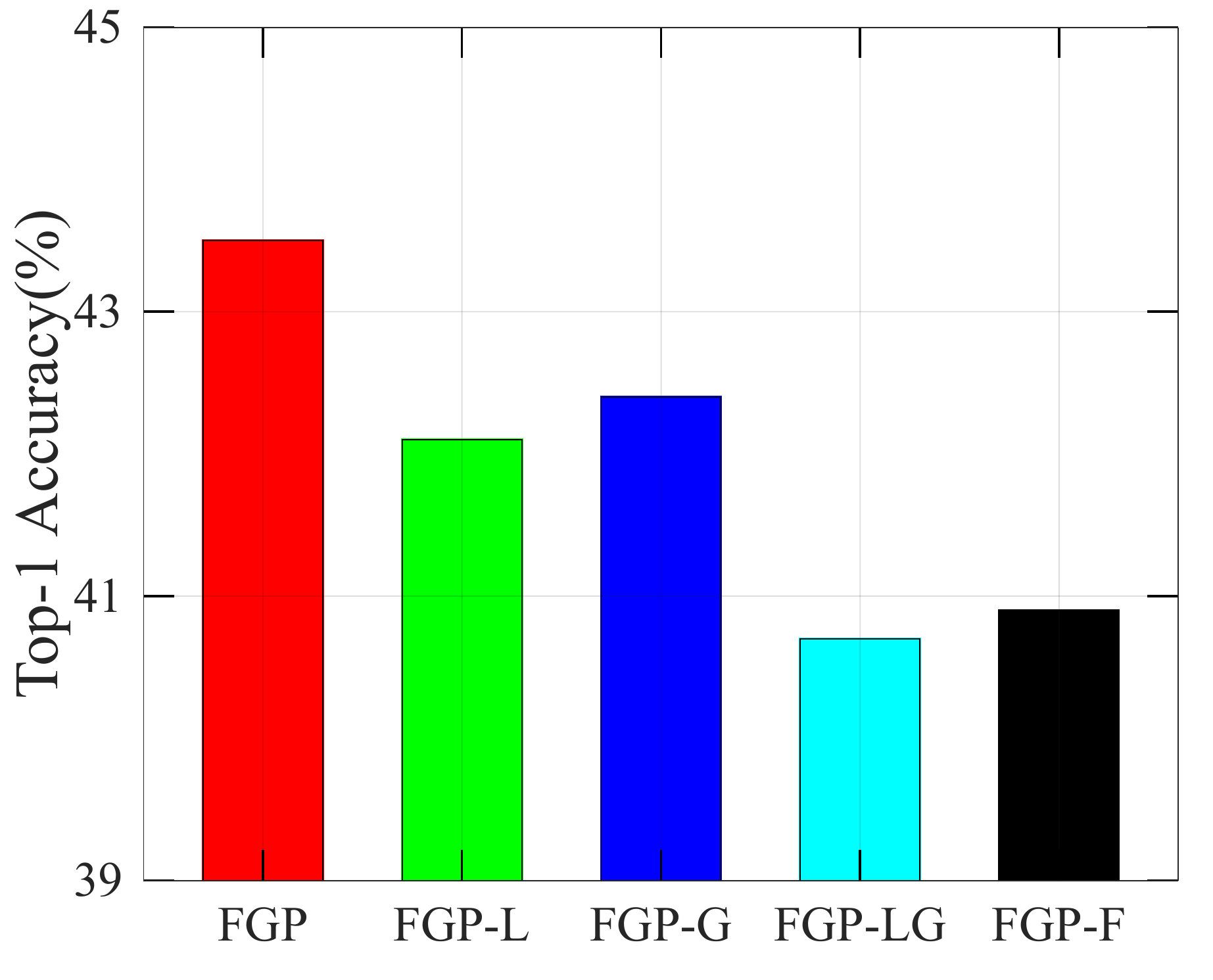}}
    % \vspace{-0.2cm}
  \caption{The module analysis of ablation studies of our proposed model FGP for both the ZSL and GZSL tasks on the ``All'' test set and the AWA2 set. -L, -G, -LG, -F indicates FGP w/o the self-calibration loss $\mathcal{L}_{sc}$, w/o the visual global features $e_g$, w/o $\mathcal{L}_{sc}$ and $e_g$, and only uses visual global features.}
  \label{fig:ablation} %% label for entire figure
  % \vspace{-0.1cm}
\end{figure}

\begin{table}[tbp]
\def\arraystretch{0.8}
\setlength{\tabcolsep}{4pt}
\large
      \caption[t]{Results of the fine-tuning experiments on the ``All'' dataset. (-f1), (-f2), and (-f1f2) indicate the FGP model without the first fine-tuning, the second fine-tuning and both fine-tunings, respectively. We also present the results of DGP and DGP(-f) (without the fine-tuning) for comparison.}
      \label{tab:resultsAblation}
      \centering
        \resizebox{.4\textwidth}{!}{ \begin{tabular}{l|c|ccccc}%c}
        %\cmidrule[1.5pt]{1-3}
        \toprule
        \bf \multirow{2}{*}{Test set} & \bf\multirow{2}{*}{Model} & \multicolumn{5}{c}{\bf Hit@k (\%)}\\
        & & 1 & 2 & 5 & 10 & 20\\
        \midrule
        {\multirow{6}{*}{\bf All}} & DGP (-f) & 2.6 & 4.7 & 9.0 & 13.5 & 19.1\\
        & DGP & 3.0 & 5.0 & 9.3 & 13.9 & 19.8\\
        \cline{2-7}
        \Tstrut & FGP (-f1) & 2.6 & 4.5 & 8.7 & 13.0 & 18.4\\
        & FGP (-f2) & 3.2 & 5.2 & 9.6 & 13.7 & 19.6\\
        & FGP (-f1f2) & 2.3 & 4.3 & 8.7 & 13.2 & 17.7\\
        & FGP & \textbf{3.7} & \textbf{5.5} & \textbf{10.5} & \textbf{14.7} & \textbf{20.9}\\
        \bottomrule
        \end{tabular}
        }
        % \vspace{-0.2cm}
\end{table}

\begin{figure}[t]
\centering
\includegraphics[width=0.32 \textwidth]{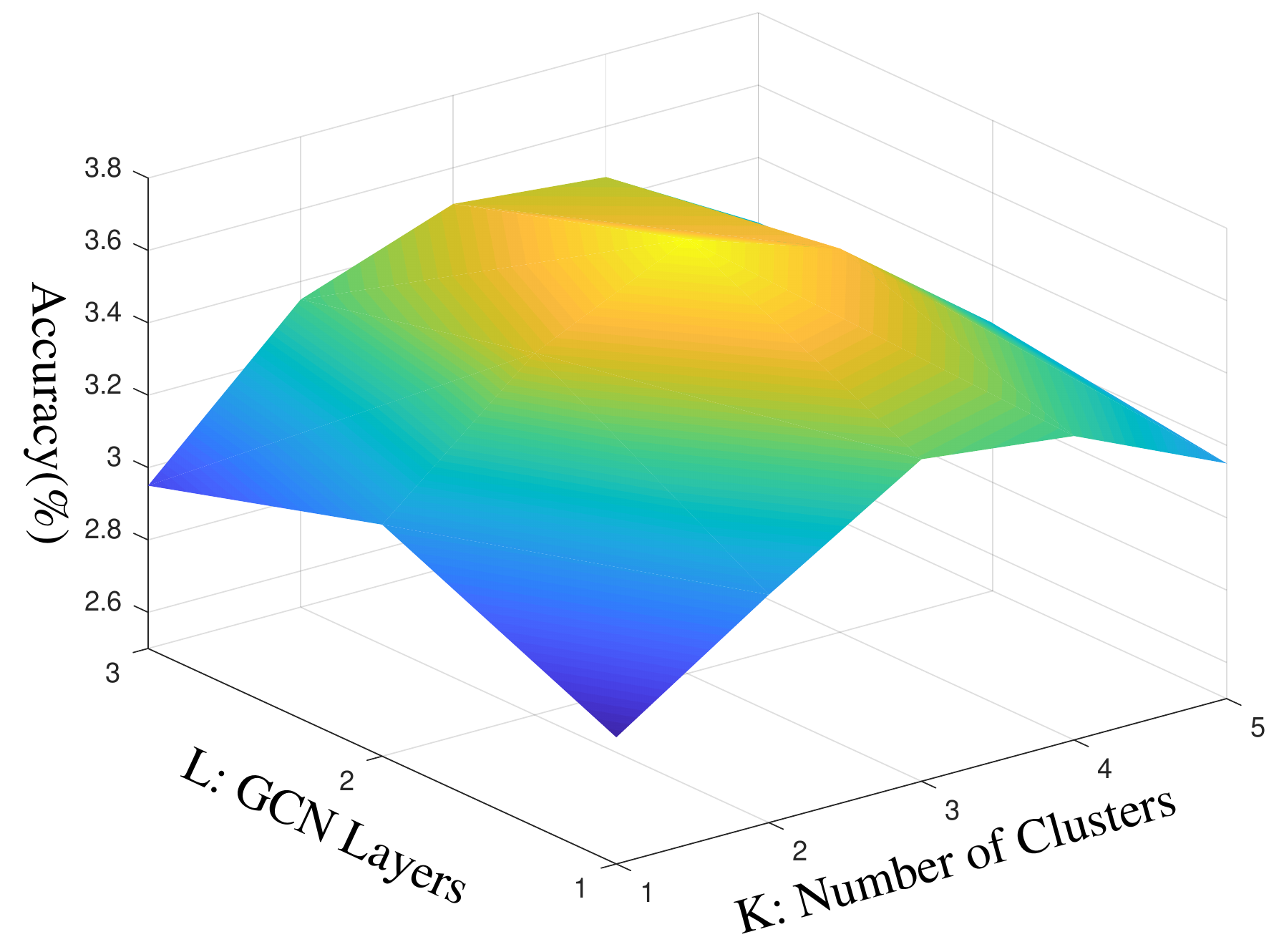}
% \vspace{-0.15cm}
\caption{The visualization results of our hyperparameter tune-up experiment for the number of GCN layers and semantic clusters on the ``All'' dataset.}
\label{fig:hp_tune}
% \vspace{-0.3cm}
\end{figure}
\subsection{Ablation Studies}

\subsubsection{Module Analysis}
To further verify the effectiveness of each modified module in our model, we design an ablation experiment carefully. In detail, the experiment includes FGP (our proposed complete model), FGP-L (FGP w/o the self-calibration loss $\mathcal{L}_{sc}$), FGP-G (FGP w/o the visual global features $\mathbf{e}_g$), FGP-LG (FGP w/o $\mathcal{L}_{sc}$ and the visual global features $\mathbf{e}_g$), and FGP-F (the model only uses visual global features, it can be viewed as DGP). We run these models on the ImageNet 2012 dataset and test their ZSL and GZSL performances both on the ``All'' set and AWA2 set, relevant experimental results are all recorded in Figure~\ref{fig:ablation}. Figure~5(a)-(b) offer the Top-1$\sim$20 accuracy on ``All''. Figure~5(c)-(d) offer the Top-1 accuracy on AWA2. We can find that the complete model FGP achieves the best results in both two tasks, but there are no distinct performance margins among other sub-models. The worst performer is FGP-LG. Due to the lack of corresponding modules, the effects of FGP-L, FGP-G, and FGP-F are obviously worse, which shows that our self-calibration loss, global visual features, and fine-grained features are all necessary and useful in the framework.
\subsubsection{Fine-tuning Analysis}
There are 2 times of fine-tuning for model training in our proposed scheme, which can be easily distinguished by reviewing \textbf{Algorithm} 1. The first fine-tuning in step 5 is used for optimizing the dense attention network and the second fine-tuning in step 8 can correct deviations between the feature extractor (including dense attention) and the replaced classifier weights learned by Multi-GCN.
We observe that fine-tuning consistently improves performance for both models in all our experiments. Ablation studies that highlight the impact of fine-tuning for the ``All'' scenario can be found in Table~\ref{tab:resultsAblation}. 
FGP (-f1f2) is viewed as the accuracy that is achieved after training the FGP model without both fine-tunings. FGP (-f1) and FGP (-f2) is used to denote the results for FGP without the first fine-tuning and the second fine-tuning, respectively. We further report the accuracy achieved by the DGP model without fine-tuning (DGP(-f)) since DGP is the base approach of FGP. We observe that the effect of FGP (-f1) and FGP (-f1f2) is much lower than FGP (-f2) and the complete model FGP, that is because the parameters of dense attention module cannot be optimized within their training phrase. Overall, fine-tuning the zero-shot learning model consistently leads to improved results.

\subsection{Hyperparameter Tune-up}
In our model, there are two major hyperparameters: GCN layers $L$ and the number of semantic clusters $K$, which affect the neighbor hops of nodes in KG and the fine-grained level respectively. We intend to explore corresponding influences bring back by these two hyperparameters in this subsection. As shown in Figure~\ref{fig:hp_tune}, the Top-1 accuracy of the FGP model is plotted in a 3D chart with different hyperparameters on the ``All'' dataset, where the value ranges of $L$ and $K$ are $[1,2,3]$ and $[1,2,3,4,5]$, respectively. In this figure, the warmer the tone, the higher the accuracy. Clearly, our model ascends to the peak when $L=2$ and $K=3$. The too-small value of $K$ leads to more overlapping of semantic space, and too-large $K$ would bring too many model parameters that result in over-fitting. Cater to the structural characteristics of the dataset, $K=3$ makes the model achieve relatively better performances than other situations, which forms the ridge of our result distribution. So the optimal number of semantic clusters is 3 in this dataset.  However, GCN layers don't show an obvious effect with the regular pattern on the performance of FGP, except for the lowest score when $K=1$. It shows that the idea of only aggregating features of entities with long paths (more neighbors) in ordinary KG has great limitations in our zero-shot learning framework.

%FGP
%FGP w/o sc-loss
%FGP w/o global-feature
%FGP w/o sc-loss global-feature
%DGP

\section{Conclusion}
In this paper, we proposed a novel Fine-grained Graph Propagation (FGP) model based on the multimodal KG, which enhanced the semantic transfer process at a fine-grained knowledge level for the zero-shot recognition task. 
%For the large-scale data without annotating attributes, it was also crucial to satisfy the visual-semantic information alignment and matching if we try to utilize the fine-grained technology.  
More specifically, we first generated key semantic embeddings which matched with relevant regions of input images in the fine-grained level visual-semantic alignment process. Then a Multi-GCN was used to learn the classifiers that mapped into all seen and unseen classes.
Along this line, after a slight fine-tuning, our model achieved the goal of embedding fine-grained features in the multimodal KG-based zero-shot learning framework.
We conducted extensive experiments on large real-world datasets and the results clearly demonstrated the effectiveness of FGP compared with several state-of-the-art methods. 
% The visualization experiment of fine-grained textual and visual features showed that our model was capable of enhancing key regions corresponding to the semantic embedding, which promoted the recognition ability for some similar categories. 
Hence our learning paradigm can be viewed as an effective booster to achieve the breakthrough of related applications which face the dilemma of sparse manual labels. The further consideration of additional knowledge graph resources such as Wikipedia for settings where these are available for a subset of nodes is a future direction. 

\section{Acknowledgments}
This research was partially supported by grants from National Key Research and Development Program of China (Grant No. 2021YFF090\\1003), National Natural Science Foundation of China (Grants No. U20A20229, 62206155, 72101176), and China Postdoctoral Science Foundation (No. 2022M720077).

% \section{Appendices}

% If your work needs an appendix, add it before the
% ``\verb|\end{document}|'' command at the conclusion of your source
% document.

% Start the appendix with the ``\verb|appendix|'' command:
% \begin{verbatim}
%   \appendix
% \end{verbatim}
% and note that in the appendix, sections are lettered, not
% numbered. This document has two appendices, demonstrating the section
% and subsection identification method.

%%
%% The next two lines define the bibliography style to be used, and
%% the bibliography file.
\bibliographystyle{ACM-Reference-Format}
% \balance
\bibliography{FDGP}

\appendix

\section{Appendices}
\begin{figure*}[t]
\centering
\includegraphics[width=0.85\textwidth]{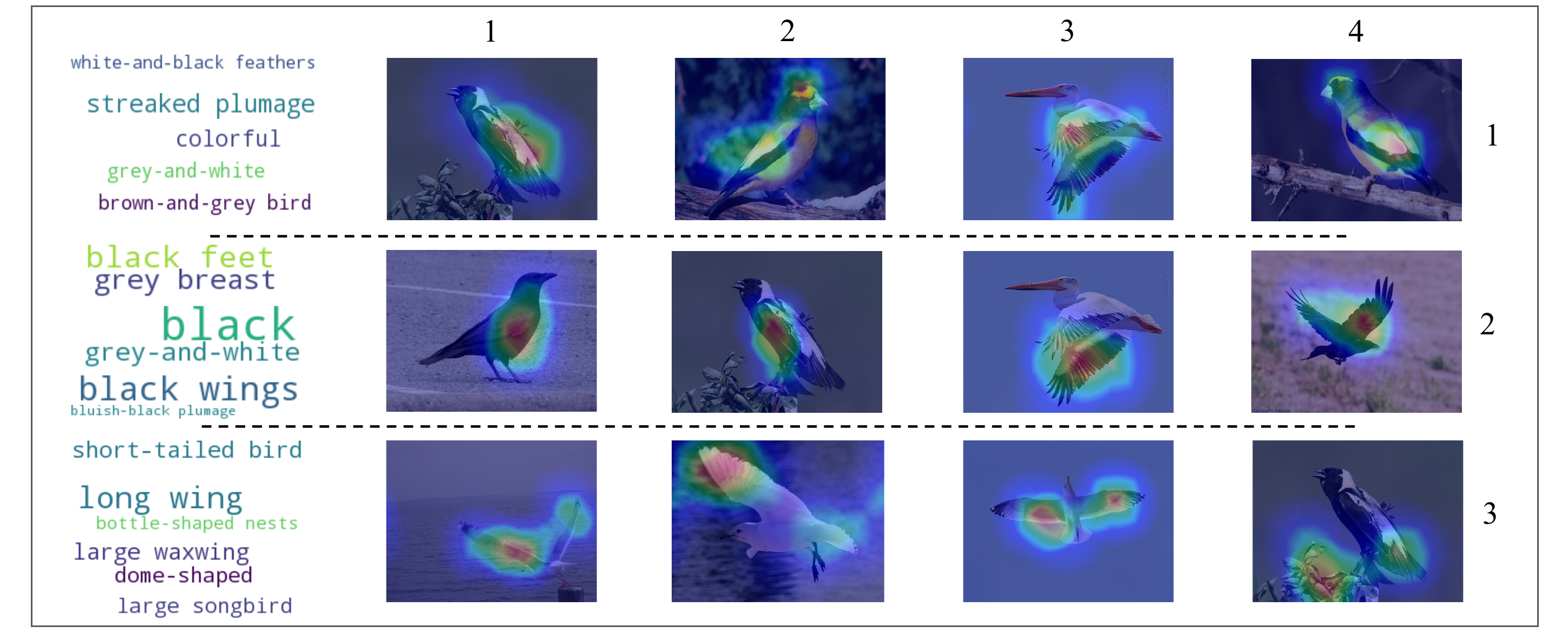}
% \vspace{-0.1cm}
\caption{The visualization results of attention maps with corresponding semantic embeddings on the bird subset. The most representative words of each semantic cluster are presented in the form of word cloud. The attention distribution of each image is clearly displayed through the heat map. }
\label{fig:case}
% \vspace{-0.3cm}
\end{figure*}
\subsection{Fine-Grained Features Visualization}
We intend to observe the focused topics of key semantic embeddings and the status of fine-grained level visual-semantic alignment. To present a more intuitive and concise qualitative result, we conduct this experiment on the collected bird subset from ImageNet which has relatively unified and distinct strong-visibility features (e.g. body, wings, feet, etc.). We show the semantic phrase clusters of our Kmeans module (here generating three clusters in our setting which are represented by the word clouds in Figure~\ref{fig:case}, the font size of a phrase denotes its centrality weight in a cluster) and corresponding cases (the images with attention heatmaps). According to Figure~\ref{fig:case}, there is a major focused topic in each semantic word cloud, and the attention heatmaps of their cases are able to match them to a considerable extent. 
For instance, keeping up with respective semantic contents, the images in the first row can strengthen the streaked and mixed-color parts, while the second row concentrates on the solid dark regions. We can find that the major focused semantics of the word clouds of the first two rows are ``streaked, colorful'' and the dark tone like ``black, grey'' too. Compared with the formers, the semantic topic of the third row does not seem very clear and obvious, so its case performances are not as good as the first two. The case$_{3,4}$ assigns some larger weights in unimportant areas, but case$_{3,1}$, case$_{3,2}$, and case$_{3,3}$ are all able to find the regions in original images that correspond to the important ``long/large wing'' semantic information. To sum up, the visualization shows that our model can realize the motivation of fine-grained feature alignment in the KG-based zero-shot learning, which helps the cognitive system distinguish similar but locally different objects more precisely such as case$_{1,1}$ and case$_{2,1}$ in Figure~\ref{fig:case}.

\end{document}